\documentclass[11pt,a4paper]{article}
\usepackage[hyperref]{emnlp2018}
\usepackage{times}
\usepackage{latexsym}
\usepackage{epsfig}
\usepackage{graphicx}

\usepackage{url}

\usepackage{amsmath}
\usepackage{amssymb}

\usepackage[linesnumbered,vlined,ruled,norelsize]{algorithm2e}

\aclfinalcopy 
\def\aclpaperid{22} 

\title{Improving the results of string kernels in sentiment analysis and Arabic dialect identification by adapting them to your test set}

\author{\hspace{7cm}Radu Tudor Ionescu$^{1,2}$ \and Andrei M. Butnaru$^1$\\
  \\
  $^1$University of Bucharest,\\
  Department of Computer Science\\
  14 Academiei, Bucharest, Romania\\
  {\tt butnaruandreimadalin@gmail.com}
 \And
  \\
  \\
  \hspace{1.5cm}$^2$Inception Institute\\
  \hspace{1.5cm}of Artificial Intelligence (IIAI)\\
  \hspace{1.5cm}Al Maryah Island, Abu Dhabi, UAE\\
  \hspace{1.5cm}{\tt raducu.ionescu@gmail.com}\\
}

\date{}

\begin{document}
\maketitle
\begin{abstract}
Recently, string kernels have obtained state-of-the-art results in various text classification tasks such as Arabic dialect identification or native language identification. In this paper, we apply two simple yet effective transductive learning approaches to further improve the results of string kernels. The first approach is based on interpreting the pairwise string kernel similarities between samples in the training set and samples in the test set as features. Our second approach is a simple self-training method based on two learning iterations. In the first iteration, a classifier is trained on the training set and tested on the test set, as usual. In the second iteration, a number of test samples (to which the classifier associated higher confidence scores) are added to the training set for another round of training. However, the ground-truth labels of the added test samples are not necessary. Instead, we use the labels predicted by the classifier in the first training iteration. By adapting string kernels to the test set, we report significantly better accuracy rates in English polarity classification and Arabic dialect identification.
\end{abstract}

\setlength{\abovedisplayskip}{3pt}
\setlength{\belowdisplayskip}{3pt}

\vspace*{-0.5cm} 
\section{Introduction}

In recent years, methods based on string kernels have demonstrated remarkable performance in various text classification tasks ranging from authorship identification \cite{PopescuG12} and sentiment analysis \cite{franco-EACL-2017,marius-KES-2017} to native language identification \cite{popescu-ionescu:2013:BEA8,ionescu-popescu-cahill-EMNLP-2014,ionescu-popescu-cahill-COLI-2016,Ionescu-BEA-2017}, dialect identification \cite{Ionescu-VarDial-2016,Radu-Andrei-ADI-2017,Ionescu-VarDial-2018} and automatic essay scoring \cite{Cozma-ACL-2018}. As long as a labeled training set is available, string kernels can reach state-of-the-art results in various languages including English \cite{ionescu-popescu-cahill-EMNLP-2014,franco-EACL-2017,Cozma-ACL-2018}, Arabic \cite{Radu-ICONIP-2015,ionescu-popescu-cahill-COLI-2016,Radu-Andrei-ADI-2017,Ionescu-VarDial-2018}, Chinese \cite{marius-KES-2017} and Norwegian \cite{ionescu-popescu-cahill-COLI-2016}. Different from all these recent approaches, we use unlabeled data from the test set to significantly increase the performance of string kernels. More precisely, we propose two transductive learning approaches combined into a unified framework. We show that the proposed framework improves the results of string kernels in two different tasks (cross-domain sentiment classification and Arabic dialect identification) and two different languages (English and Arabic). To the best of our knowledge, transductive learning frameworks based on string kernels have not been studied in previous works.

\vspace*{-0.2cm} 
\section{Transductive String Kernels}
\label{sec_String_Kernels}
\vspace*{-0.1cm} 

\begin{figure}[!t]
\begin{center}
\includegraphics[width=1.0\linewidth]{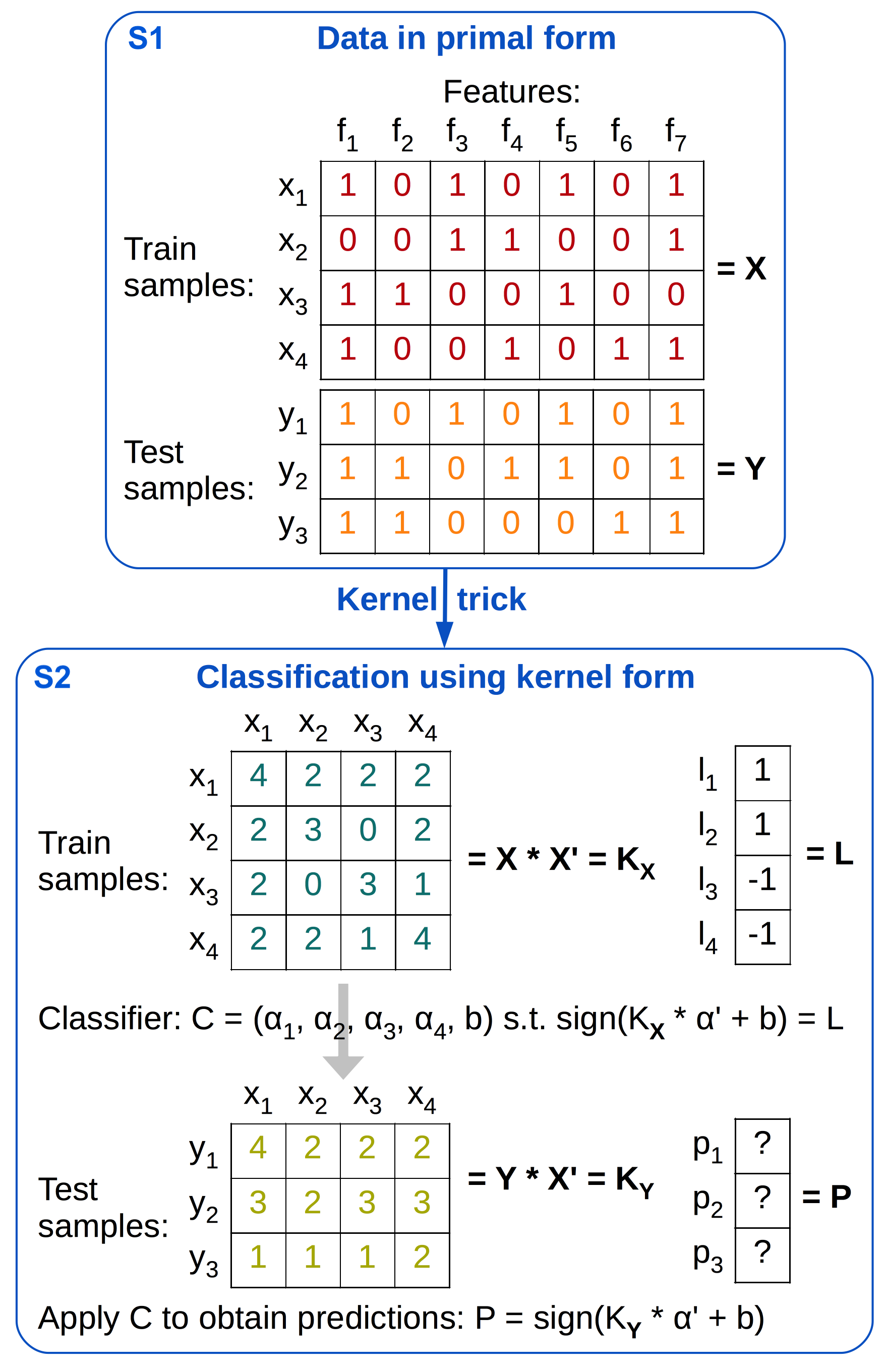}
\end{center}
\vspace*{-0.6cm}
\caption{The standard kernel learning pipeline based on the linear kernel. Kernel normalization is not illustrated for simplicity. Best viewed in color.}
\label{fig1}
\vspace*{-0.5cm}
\end{figure}

\noindent
{\bf String kernels.}
Kernel functions \cite{taylor-Cristianini-cup-2004} capture the intuitive notion of similarity between objects in a specific domain. For example, in text mining, string kernels can be used to measure the pairwise similarity between text samples, simply based on character n-grams. Various string kernel functions have been proposed to date \cite{LodhiSSCW02,taylor-Cristianini-cup-2004,ionescu-popescu-cahill-EMNLP-2014}. Perhaps one of the most recently introduced string kernels is the histogram intersection string kernel \cite{ionescu-popescu-cahill-EMNLP-2014}. For two strings over an alphabet $\Sigma$, $x,y \in \Sigma^*$, the intersection string kernel is formally defined as follows:
\begin{equation}
\begin{split}
k^{\cap}(x,y)=\sum\limits_{v \in \Sigma^p} \min \lbrace \mbox{num}_v(x), \mbox{num}_v(y) \rbrace ,
\end{split}
\end{equation}
where $\mbox{num}_v(x)$ is the number of occurrences of n-gram $v$ as a substring in $x$, and $p$ is the length of $v$. The spectrum string kernel or the presence bits string kernel can be defined in a similar fashion \cite{ionescu-popescu-cahill-EMNLP-2014}. The standard kernel learning pipeline is presented in Figure~\ref{fig1}. String kernels help to efficiently~\cite{marius-KES-2017} compute the dual representation directly, thus skipping the first step in the pipeline illustrated in Figure~\ref{fig1}.

\begin{figure}[!th]
\begin{center}
\includegraphics[width=1.0\linewidth]{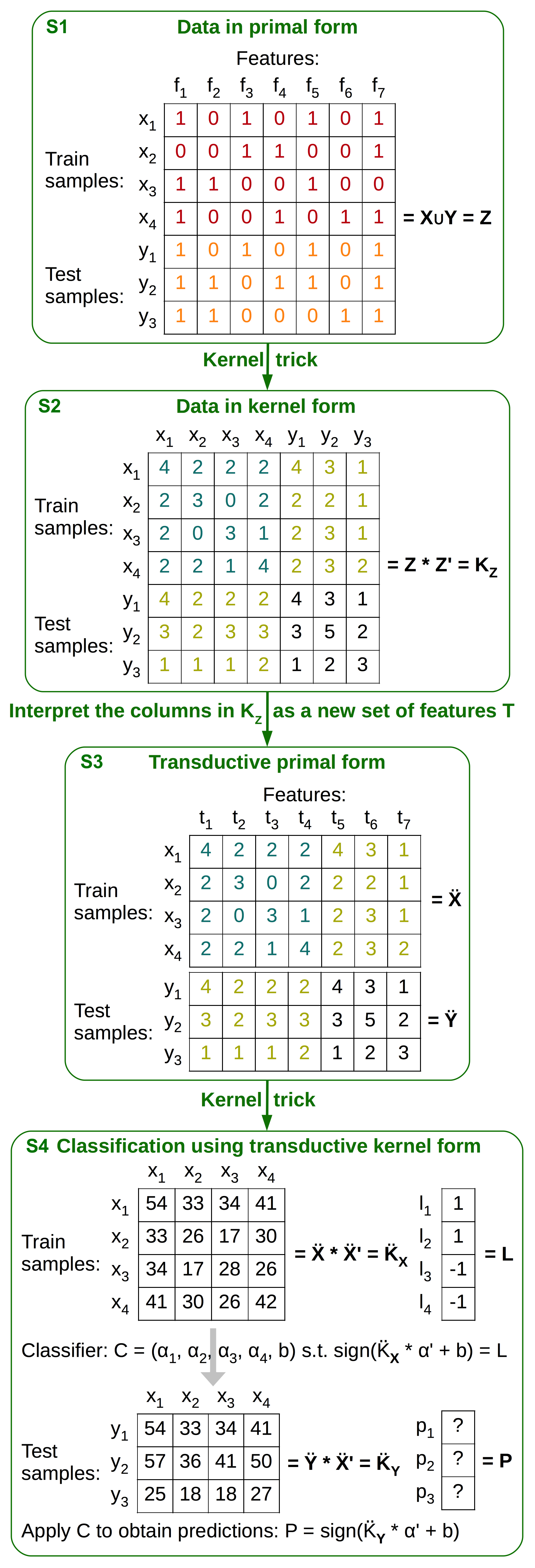}
\end{center}
\vspace*{-0.6cm}
\caption{The transductive kernel learning pipeline based on the linear kernel. Kernel normalization and RBF kernel transformation are not illustrated for simplicity. Best viewed in color.}
\label{fig2}
\vspace*{-0.5cm}
\end{figure}

\noindent
{\bf Transductive string kernels.}
We propose a simple and straightforward approach to produce a transductive similarity measure suitable for strings, as illustrated in Figure~\ref{fig2}. We take the following steps to derive transductive string kernels. For a given kernel (similarity) function $k$, we first build the full kernel matrix $K$, by including the pairwise similarities of samples from both the train and the test sets (step $S1$ in Figure~\ref{fig2}) . For a training set $X = \{x_1, x_2, ..., x_m\}$ of $m$ samples and a test set $Y = \{y_1, y_2, ..., y_n\}$ of $n$ samples, such that $X \cap Y = \emptyset$, each component in the full kernel matrix is defined as follows (step $S2$ in Figure~\ref{fig2}):
\begin{equation}
\begin{split}
K_{ij}= k(z_i, z_j),
\end{split}
\end{equation}
where $z_i$ and $z_j$ are samples from the set $Z = X \cup Y = \{x_1, x_2, ..., x_m, y_1, y_2, ..., y_n\}$, for all $1 \leq i,j \leq m + n$. We then normalize the kernel matrix by dividing each component by the square root of the product of the two corresponding diagonal components:
\begin{equation}\label{eq_Kernel_Matrix_Normalization}
\begin{split}
\hat{K}_{ij} = \frac{K_{ij}}{\sqrt{K_{ii} \cdot K_{jj}}}.
\end{split}
\end{equation}

We transform the normalized kernel matrix into a radial basis function (RBF) kernel matrix as follows:
\begin{equation}\label{eq_RBF_Kernel}
\begin{split}
\tilde{K}_{ij} = exp \left( - \frac{\displaystyle 1 - \hat{K}_{ij}} {\displaystyle 2 \sigma^2} \right).
\end{split}
\end{equation}
As the kernel matrix is already normalized, we can choose $\sigma^2 = 0.5$ for simplicity. Therefore, Equation~\eqref{eq_RBF_Kernel} becomes:
\begin{equation}\label{eq_RBF_Kernel_simple}
\begin{split}
\tilde{K}_{ij} = exp \left(-1 + \hat{K}_{ij}\right).
\end{split}
\end{equation}
Each row in the RBF kernel matrix $\tilde{K}$ is now interpreted as a feature vector, going from step $S2$ to step $S3$ in Figure~\ref{fig2}. In other words, each sample $z_i$ is represented by a feature vector that contains the similarity between the respective sample $z_i$ and all the samples in $Z$ (step $S3$ in Figure~\ref{fig2}). Since $Z$ includes the test samples as well, the feature vector is inherently adapted to the test set. Indeed, it is easy to see that the features will be different if we choose to apply the string kernel approach on a set of test samples $Y'$, such that $Y' \neq Y$. It is important to note that through the features, the subsequent classifier will have some information about the test samples at training time. More specifically, the feature vector conveys information about how similar is every test sample to every training sample.
We next consider the linear kernel, which is given by the scalar product between the new feature vectors. To obtain the final linear kernel matrix, we simply need to compute the product between the RBF kernel matrix and its transpose (step $S4$ in Figure~\ref{fig2}):
\begin{equation}
\ddot{K} = \tilde{K} \cdot \tilde{K}'.
\end{equation}
In this way, the samples from the test set, which are included in $Z$, are used to obtain new (transductive) string kernels that are adapted to the test set at hand.

\noindent
{\bf Transductive kernel classifier.}
After obtaining the transductive string kernels, we use a simple transductive learning approach that falls in the category of self-training methods \cite{McClosky-NAACL-2006,Chen-NIPS-2011}. The transductive approach is divided into two learning iterations. In the first iteration, a kernel classifier is trained on the training data and applied on the test data, just as usual. 
Next, the test samples are sorted by the classifier's confidence score to maximize the probability of correctly predicted labels in the top of the sorted list. In the second iteration, a fixed number of samples ($1000$ in the experiments) from the top of the list are added to the training set for another round of training. Even though a small percent (less than $8\%$ in all experiments) of the predicted labels corresponding to the newly included samples are wrong, the classifier has the chance to learn some useful patterns (from the correctly predicted labels) only visible in the test data. The transductive kernel classifier (TKC) is based on the intuition that the added test samples bring more useful information than noise, since the majority of added test samples have correct labels. Finally, we would like to stress out that \emph{the ground-truth test labels are never used in our transductive algorithm}.

The proposed transductive learning approaches are used together in a unified framework. As any other transductive learning method, the main disadvantage of the proposed framework is that the \emph{unlabeled} test samples from the target domain need to be used in the training stage. Nevertheless, we present empirical results indicating that our approach can obtain \emph{significantly better} accuracy rates in cross-domain polarity classification and Arabic dialect identification compared to state-of-the-art methods based on string kernels \cite{franco-EACL-2017,Radu-Andrei-ADI-2017}. We also report better results than other domain adaptation methods \cite{Pan-WWW-2010,Bollegala-KDE-2013,Franco-KBS-2015,Sun-AAAI-2016,Huang-AAAI-2017}.

\vspace*{-0.2cm} 
\section{Polarity Classification}
\label{sec_Polarity_Experiments}
\vspace*{-0.1cm} 

\noindent
{\bf Data set.}
For the cross-domain polarity classification experiments, we use the second version of Multi-Domain Sentiment Dataset \cite{Blitzer-ACL-2007}. The data set contains Amazon product reviews of four different domains: Books (B), DVDs (D), Electronics (E) and Kitchen appliances (K). Reviews contain star ratings (from 1 to 5) which are converted into binary labels as follows: reviews rated with more than 3 stars are labeled as positive, and those with less than 3 stars as negative. In each domain, there are 1000 positive and 1000 negative reviews.

\noindent
{\bf Baselines.}
We compare our approach with several methods \cite{Pan-WWW-2010,Bollegala-KDE-2013,Franco-KBS-2015,Sun-AAAI-2016,franco-EACL-2017,Huang-AAAI-2017} in two cross-domain settings. Using string kernels, \newcite{franco-EACL-2017} reported better performance than SST \cite{Bollegala-KDE-2013} and KE-Meta \cite {Franco-KBS-2015} in the multi-source domain setting. In addition, we compare our approach with SFA \cite{Pan-WWW-2010}, KMM \cite{Huang-NIPS-2007}, CORAL \cite{Sun-AAAI-2016} and TR-TrAdaBoost \cite{Huang-AAAI-2017} in the single-source setting. 

\begin{table}[!t]
\setlength\tabcolsep{3.5pt}
\small{
\begin{center}
\begin{tabular}{lcccc}
\hline
Method 			& DEK$\rightarrow$B	&	BEK$\rightarrow$D	&	BDK$\rightarrow$E	& BDE$\rightarrow$K\\
\hline
SST										& $76.3$						& $78.3$						& $83.9$						& $85.2$ \\
KE-Meta     							& $77.9$						& $80.4$						& $78.9$						& $82.5$ \\
$K_{0/1}$ (sota)				& $82.0$ 					& $81.9$ 					& $83.6$ 					& $85.1$ \\
$K_{\cap}$ (sota)				& $80.7$ 					& $80.7$ 					& $83.0$ 					& $85.2$ \\
\hline
\vspace{-0.9em}\\
$\ddot{K}_{0/1}$				& $82.9$						& $83.2$*					& $84.8$*					& $86.0$*\\
$\ddot{K}_{\cap}$				& $82.5$						& $82.9$*					& $84.5$*					& $86.1$*\\
$\ddot{K}_{0/1}$ + TKC 	& $\mathbf{84.1}$* 	& $\mathbf{84.0}$* 	& $\mathbf{85.4}$* 	& $86.9$*\\
$\ddot{K}_{\cap}$ + TKC	& $83.8$*					& $83.5$*					& $85.0$*					& $\mathbf{87.1}$*\\
\hline
\end{tabular}
\end{center}
}
\vspace*{-0.3cm}
\caption{Multi-source cross-domain polarity classification accuracy rates (in $\%$) of our transductive approaches versus a state-of-the-art (sota) baseline based on string kernels \cite{franco-EACL-2017}, as well as SST \cite{Bollegala-KDE-2013} and KE-Meta \cite{Franco-KBS-2015}. The best accuracy rates are highlighted in bold. The marker * indicates that the performance is significantly better than the best baseline string kernel according to a paired McNemar's test performed at a significance level of $0.01$.}
\label{tab_Polarity_Multi}
\vspace*{-0.5cm}
\end{table}

\noindent
{\bf Evaluation procedure and parameters.}
We follow the same evaluation methodology of \newcite{franco-EACL-2017}, to ensure a fair comparison. Furthermore, we use the same kernels, namely the presence bits string kernel ($K_{0/1}$) and the intersection string kernel ($K_{\cap}$), and the same range of character n-grams (5-8). To compute the string kernels, we used the open-source code provided by \newcite{radu-marius-book-chap6-2016}.
For the transductive kernel classifier, we select $r=1000$ unlabeled test samples to be included in the training set for the second round of training. We choose Kernel Ridge Regression \cite{taylor-Cristianini-cup-2004} as classifier and set its regularization parameter to $10^{-5}$ in all our experiments. Although \newcite{franco-EACL-2017} used a different classifier, namely Kernel Discriminant Analysis, we observed that Kernel Ridge Regression produces similar results ($\pm 0.1\%$) when we employ the same string kernels.
As \newcite{franco-EACL-2017}, we evaluate our approach in two cross-domain settings. In the multi-source setting, we train the models on all domains, except the one used for testing. In the single-source setting, we train the models on one of the four domains and we independently test the models on the remaining three domains.

\begin{table*}[!th]
\setlength\tabcolsep{4.5pt}
\small{
\begin{center}
\begin{tabular}{lcccccccccccc}
\hline
Method 												& D$\rightarrow$B			& E$\rightarrow$B		& K$\rightarrow$B		
															& B$\rightarrow$D 			& E$\rightarrow$D 		& K$\rightarrow$D
															& B$\rightarrow$E 			& D$\rightarrow$E 		& K$\rightarrow$E
															& B$\rightarrow$K 			& D$\rightarrow$K 		& E$\rightarrow$K\\
\hline
SFA														& $79.8$ 						& $78.3$ 					& $75.2$ 
															& $81.4$ 						& $77.2$ 					& $\mathbf{78.5}$
															& $73.5$ 						& $76.7$ 					& $85.1$ 
															& $79.1$ 						& $80.8$ 					& $86.8$\\

KMM														& $78.6$ 						& - 							& - 
															& - 								& - 							& $72.2$
															& $\mathbf{76.9}$ 		& - 							& - 
															& - 								& - 							& $83.6$\\
																														
CORAL													& $78.3$ 						& - 							& - 
															& - 								& - 							& $73.9$
															& $76.3$ 						& - 							& - 
															& - 								& - 							& $83.6$\\
															
TR-TrAdaBoost										& $74.7$ 						& $69.1$ 					& $70.6$ 
															& $79.6$ 						& $71.8$ 					& $74.4$
															& $74.9$ 						& $75.9$ 					& $83.1$ 
															& $77.8$ 						& $75.7$ 					& $83.7$\\
															
$K_{0/1}$ (sota)									& $82.0$ 						& $72.4$ 					& $72.7$ 
															& $81.4$ 						& $74.9$ 					& $73.6$
															& $71.3$ 						& $74.4$ 					& $83.9$ 
															& $74.6$ 						& $75.4$ 					& $84.9$\\
																
$K_{\cap}$ 	(sota)								& $82.1$ 						& $72.4$ 					& $72.8$ 
															& $81.3$ 						& $75.1$					& $72.9$
															& $71.8$ 						& $74.5$ 					& $84.4$
															& $74.9$ 						& $75.1$ 					& $84.9$\\
\hline
\vspace{-0.9em}\\
$\ddot{K}_{0/1}$									& $83.3$* 					& $74.5$* 					& $74.3$*
															& $83.0$* 					& $76.9$* 					& $74.9$*
															& $74.0$* 					& $76.0$* 					& $85.4$*
															& $77.6$* 					& $77.3$* 					& $86.0$*\\
																					
$\ddot{K}_{\cap}$									& $83.2$* 					& $74.2$* 					& $74.0$*
															& $82.8$* 					& $76.4$* 					& $75.1$*
															& $74.2$* 					& $75.9$* 					& $85.2$*
															& $77.6$* 					& $77.3$* 					& $85.9$*\\
																					
$\ddot{K}_{0/1}$ + TKC 						& $\mathbf{84.9}$* 		& $\mathbf{78.5}$* 		& $\mathbf{76.6}$*
															& $84.0$* 					& $\mathbf{79.6}$* 		& $76.4$*
															& $76.6$* 					& $\mathbf{77.1}$* 		& $\mathbf{86.4}$*
															& $\mathbf{79.6}$* 		& $\mathbf{80.9}$* 		& $\mathbf{87.0}$*\\
																					
$\ddot{K}_{\cap}$ + TKC						& $84.5$* 					& $\mathbf{78.5}$* 		& $75.8$*				
															& $\mathbf{84.2}$* 		& $79.1$* 					& $76.5$*
															& $76.7$* 					& $76.8$* 					& $\mathbf{86.4}$*
															& $79.4$* 					& $80.5$* 					& $\mathbf{87.0}$*\\
\hline
\end{tabular}
\end{center}
}
\vspace*{-0.3cm}
\caption{Single-source cross-domain polarity classification accuracy rates (in $\%$) of our transductive approaches versus a state-of-the-art (sota) baseline based on string kernels \cite{franco-EACL-2017}, as well as SFA \cite{Pan-WWW-2010}, KMM \cite{Huang-NIPS-2007}, CORAL \cite{Sun-AAAI-2016} and TR-TrAdaBoost \cite{Huang-AAAI-2017}. The best accuracy rates are highlighted in bold. The marker * indicates that the performance is significantly better than the best baseline string kernel according to a paired McNemar's test performed at a significance level of $0.01$.}
\label{tab_Polarity_Single}
\vspace*{-0.5cm}
\end{table*}

\noindent
{\bf Results in multi-source setting.}
The results for the multi-source cross-domain polarity classification setting are presented in Table~\ref{tab_Polarity_Multi}. Both the transductive presence bits string kernel ($\ddot{K}_{0/1}$) and the transductive intersection kernel ($\ddot{K}_{\cap}$) obtain better results than their original counterparts. 
Moreover, according to the McNemar's test~\cite{Dietterich-NC-1998}, the results on the DVDs, the Electronics and the Kitchen target domains are significantly better than the best baseline string kernel, with a confidence level of $0.01$. 
When we employ the transductive kernel classifier (TKC), we obtain even better results. On all domains, the accuracy rates yielded by the transductive classifier are more than $1.5\%$ better than the best baseline. For example, on the Books domain the accuracy of the transductive classifier based on the presence bits kernel ($84.1\%$) is $2.1\%$ above the best baseline ($82.0\%$) represented by the intersection string kernel. Remarkably, the improvements brought by our transductive string kernel approach are statistically significant in all domains.

\noindent
{\bf Results in single-source setting.} The results for the single-source cross-domain polarity classification setting are presented in Table~\ref{tab_Polarity_Single}. We considered all possible combinations of source and target domains in this experiment, and we improve the results in each and every case. Without exception, the accuracy rates reached by the transductive string kernels are significantly better than the best baseline string kernel \cite{franco-EACL-2017}, according to the McNemar's test performed at a confidence level of $0.01$. The highest improvements (above $2.7\%$) are obtained when the source domain contains Books reviews and the target domain contains Kitchen reviews. 
As in the multi-source setting, we obtain much better results when the transductive classifier is employed for the learning task. In all cases, the accuracy rates of the transductive classifier are more than $2\%$ better than the best baseline string kernel. Remarkably, in four cases (E$\rightarrow$B, E$\rightarrow$D, B$\rightarrow$K and D$\rightarrow$K) our improvements are greater than $4\%$. 
The improvements brought by our transductive classifier based on string kernels are statistically significant in each and every case. In comparison with SFA \cite{Pan-WWW-2010}, we obtain better results in all but one case (K$\rightarrow$D). With respect to KMM \cite{Huang-NIPS-2007}, we also obtain better results in all but one case (B$\rightarrow$E). Remarkably, we surpass the other state-of-the-art approaches \cite{Sun-AAAI-2016,Huang-AAAI-2017} in all cases.

\vspace*{-0.2cm} 
\section{Arabic Dialect Identification}
\label{sec_Dialect_Experiments}
\vspace*{-0.1cm} 

\noindent
{\bf Data set.}
The Arabic Dialect Identification (ADI) data set \cite{Ali-2016} contains audio recordings and Automatic Speech Recognition (ASR) transcripts of Arabic speech collected from the Broadcast News domain. The classification task is to discriminate between Modern Standard Arabic and four Arabic dialects, namely Egyptian, Gulf, Levantine, and Maghrebi. The training set contains 14000 samples, the development set contains 1524 samples, and the test contains another 1492 samples. The data set was used in the ADI Shared Task of the 2017 VarDial Evaluation Campaign \cite{dsl2017}. 

\noindent
{\bf Baseline.}
We choose as baseline the approach of \newcite{Radu-Andrei-ADI-2017}, which is based on string kernels and multiple kernel learning. The approach that we consider as baseline is the winner of the 2017 ADI Shared Task  \cite{dsl2017}. 
In addition, we also compare with the second-best approach (Meta-classifier) \cite{malmasi-VarDial-2017}.

\noindent
{\bf Evaluation procedure and parameters.}
\newcite{Radu-Andrei-ADI-2017} combined four kernels into a sum, and used Kernel Ridge Regression for training. Three of the kernels are based on character n-grams extracted from ASR transcripts. These are the presence bits string kernel ($K_{0/1}$), the intersection string kernel ($K_{\cap}$), and a kernel based on Local Rank Distance ($K_{LRD}$) \cite{radu-LRD-synasc-2013}. The fourth kernel is an RBF kernel ($K_{ivec}$) based on the i-vectors provided with the ADI data set \cite{Ali-2016}. In our experiments, we employ the exact same kernels as \newcite{Radu-Andrei-ADI-2017} to ensure an unbiased comparison with their approach. 
As in the polarity classification experiments, we select $r=1000$ unlabeled test samples to be included in the training set for the second round of training the transductive classifier, and we use Kernel Ridge Regression with a regularization of $10^{-5}$ in all our ADI experiments. 

\begin{table}[!t]
\setlength\tabcolsep{3.5pt}
\small{
\begin{center}
\begin{tabular}{lcc}
\hline
Method 																								& Development			&	Test \\
\hline
Meta-classifier							
																											& -								& $71.65$\\
$K_{0/1}$+$K_{\cap}$+$K_{LRD}$+$K_{ivec}$	(sota)
																											& $64.17$ 					& $76.27$ \\
\hline
\vspace{-0.9em}\\
$\ddot{K}_{0/1}$+$\ddot{K}_{\cap}$+$\ddot{K}_{LRD}$+$\ddot{K}_{ivec}$
																											& $65.42$*					& $77.08$\\
$\ddot{K}_{0/1}$+$\ddot{K}_{\cap}$+$\ddot{K}_{LRD}$+$\ddot{K}_{ivec}$	 + TKC
																											& $\mathbf{66.73}$*	& $\mathbf{78.35}$*\\
\hline
\end{tabular}
\end{center}
}
\vspace*{-0.3cm}
\caption{Arabic dialect identification accuracy rates (in $\%$) of our adapted string kernels versus the 2017 ADI Shared Task winner (sota) \cite{Radu-Andrei-ADI-2017} and the first runner up \cite{malmasi-VarDial-2017}. The best accuracy rates are highlighted in bold. The marker * indicates that the performance is significantly better than \cite{Radu-Andrei-ADI-2017} according to a paired McNemar's test performed at a significance level of $0.01$.}
\label{tab_ADI}
\vspace*{-0.5cm}
\end{table}

\noindent
{\bf Results.}
The results for the cross-domain Arabic dialect identification experiments on both the development and the test sets are presented in Table~\ref{tab_ADI}. The domain-adapted sum of kernels obtains improvements above $0.8\%$ over the state-of-the-art sum of kernels \cite{Radu-Andrei-ADI-2017}. The improvement on the development set (from $64.17\%$ to $65.42\%$) is statistically significant. Nevertheless, we obtain higher and significant improvements when we employ the transductive classifier. Our best accuracy is $66.73\%$ ($2.56\%$ above the baseline) on the development set and $78.35\%$ ($2.08\%$ above the baseline) on the test set.
The results show that our domain adaptation framework based on string kernels attains the best performance on the ADI Shared Task data set, and the improvements over the state-of-the-art are statistically significant, according to the McNemar's test. 

\bibliography{references}

\begin{thebibliography}{29}
\expandafter\ifx\csname natexlab\endcsname\relax\def\natexlab#1{#1}\fi

\bibitem[{Ali et~al.(2016)Ali, Dehak, Cardinal, Khurana, Yella, Glass, Bell,
  and Renals}]{Ali-2016}
Ahmed Ali, Najim Dehak, Patrick Cardinal, Sameer Khurana, Sree~Harsha Yella,
  James Glass, Peter Bell, and Steve Renals. 2016.
\newblock Automatic dialect detection in arabic broadcast speech.
\newblock In \emph{Proceedings of INTERSPEECH}, pages 2934--2938.

\bibitem[{Blitzer et~al.(2007)Blitzer, Dredze, and Pereira}]{Blitzer-ACL-2007}
John Blitzer, Mark Dredze, and Fernando Pereira. 2007.
\newblock Biographies, bollywood, boomboxes and blenders: Domain adaptation for
  sentiment classification.
\newblock In \emph{Proceedings of ACL}, pages 187--205.

\bibitem[{Bollegala et~al.(2013)Bollegala, Weir, and
  Carroll}]{Bollegala-KDE-2013}
D.~Bollegala, D.~Weir, and J.~Carroll. 2013.
\newblock {Cross-Domain Sentiment Classification Using a Sentiment Sensitive
  Thesaurus}.
\newblock \emph{IEEE Transactions on Knowledge and Data Engineering},
  25(8):1719--1731.

\bibitem[{Butnaru and Ionescu(2018)}]{Ionescu-VarDial-2018}
Andrei~M. Butnaru and Radu~Tudor Ionescu. 2018.
\newblock {UnibucKernel Reloaded: First Place in Arabic Dialect Identification
  for the Second Year in a Row}.
\newblock In \emph{Proceedings of VarDial Workshop of COLING}, pages 77--87.

\bibitem[{Chen et~al.(2011)Chen, Weinberger, and Blitzer}]{Chen-NIPS-2011}
Minmin Chen, Kilian Weinberger, and John Blitzer. 2011.
\newblock {Co-Training for Domain Adaptation}.
\newblock In \emph{Proceedings of NIPS}, pages 2456--2464.

\bibitem[{Cozma et~al.(2018)Cozma, Butnaru, and Ionescu}]{Cozma-ACL-2018}
M\u{a}d\u{a}lina Cozma, Andrei Butnaru, and Radu~Tudor Ionescu. 2018.
\newblock Automated essay scoring with string kernels and word embeddings.
\newblock In \emph{Proceedings of ACL}, pages 503--509.

\bibitem[{Dietterich(1998)}]{Dietterich-NC-1998}
Thomas~G. Dietterich. 1998.
\newblock {Approximate Statistical Tests for Comparing Supervised
  Classification Learning Algorithms}.
\newblock \emph{Neural Computation}, 10(7):1895--1923.

\bibitem[{Franco-Salvador et~al.(2015)Franco-Salvador, Cruz, Troyano, and
  Rosso}]{Franco-KBS-2015}
Marc Franco-Salvador, Fermin~L. Cruz, Jose~A. Troyano, and Paolo Rosso. 2015.
\newblock Cross-domain polarity classification using a knowledge-enhanced
  meta-classifier.
\newblock \emph{Knowledge-Based Systems}, 86:46--56.

\bibitem[{Gim\'{e}nez-P\'{e}rez et~al.(2017)Gim\'{e}nez-P\'{e}rez,
  Franco-Salvador, and Rosso}]{franco-EACL-2017}
Rosa~M. Gim\'{e}nez-P\'{e}rez, Marc Franco-Salvador, and Paolo Rosso. 2017.
\newblock {Single and Cross-domain Polarity Classification using String
  Kernels}.
\newblock In \emph{Proceedings of EACL}, pages 558--563.

\bibitem[{Huang et~al.(2007)Huang, Gretton, Borgwardt, Sch{\"o}lkopf, and
  Smola}]{Huang-NIPS-2007}
Jiayuan Huang, Arthur Gretton, Karsten Borgwardt, Bernhard Sch{\"o}lkopf, and
  Alex Smola. 2007.
\newblock Correcting sample selection bias by unlabeled data.
\newblock In \emph{Proceedings of NIPS}, pages 601--608.

\bibitem[{Huang et~al.(2017)Huang, Rao, Xie, Wong, and Wang}]{Huang-AAAI-2017}
Xingchang Huang, Yanghui Rao, Haoran Xie, Tak-Lam Wong, and Fu~Lee Wang. 2017.
\newblock {Cross-Domain Sentiment Classification via Topic-Related TrAdaBoost}.
\newblock In \emph{Proceedings of AAAI}, pages 4939--4940.

\bibitem[{Ionescu(2013)}]{radu-LRD-synasc-2013}
Radu~Tudor Ionescu. 2013.
\newblock {Local Rank Distance}.
\newblock In \emph{Proceedings of SYNASC}, pages 221--228.

\bibitem[{Ionescu(2015)}]{Radu-ICONIP-2015}
Radu~Tudor Ionescu. 2015.
\newblock {A Fast Algorithm for Local Rank Distance: Application to Arabic
  Native Language Identification}.
\newblock In \emph{Proceedings of ICONIP}, volume 9490, pages 390--400.

\bibitem[{Ionescu and Butnaru(2017)}]{Radu-Andrei-ADI-2017}
Radu~Tudor Ionescu and Andrei Butnaru. 2017.
\newblock {Learning to Identify Arabic and German Dialects using Multiple
  Kernels}.
\newblock In \emph{Proceedings of VarDial Workshop of EACL}, pages 200--209.

\bibitem[{Ionescu and
  Popescu(2016{\natexlab{a}})}]{radu-marius-book-chap6-2016}
Radu~Tudor Ionescu and Marius Popescu. 2016{\natexlab{a}}.
\newblock {Native Language Identification with String Kernels}.
\newblock In \emph{{Knowledge Transfer between Computer Vision and Text
  Mining}}, Advances in Computer Vision and Pattern Recognition, chapter~8,
  pages 193--227. Springer International Publishing.

\bibitem[{Ionescu and Popescu(2016{\natexlab{b}})}]{Ionescu-VarDial-2016}
Radu~Tudor Ionescu and Marius Popescu. 2016{\natexlab{b}}.
\newblock {UnibucKernel: An Approach for Arabic Dialect Identification based on
  Multiple String Kernels}.
\newblock In \emph{Proceedings of VarDial Workshop of COLING}, pages 135--144.

\bibitem[{Ionescu and Popescu(2017)}]{Ionescu-BEA-2017}
Radu~Tudor Ionescu and Marius Popescu. 2017.
\newblock Can string kernels pass the test of time in native language
  identification?
\newblock In \emph{Proceedings of the 12th Workshop on Innovative Use of NLP
  for Building Educational Applications}, pages 224--234.

\bibitem[{Ionescu et~al.(2014)Ionescu, Popescu, and
  Cahill}]{ionescu-popescu-cahill-EMNLP-2014}
Radu~Tudor Ionescu, Marius Popescu, and Aoife Cahill. 2014.
\newblock Can characters reveal your native language? a language-independent
  approach to native language identification.
\newblock In \emph{Proceedings of EMNLP}, pages 1363--1373.

\bibitem[{Ionescu et~al.(2016)Ionescu, Popescu, and
  Cahill}]{ionescu-popescu-cahill-COLI-2016}
Radu~Tudor Ionescu, Marius Popescu, and Aoife Cahill. 2016.
\newblock String kernels for native language identification: Insights from
  behind the curtains.
\newblock \emph{Computational Linguistics}, 42(3):491--525.

\bibitem[{Lodhi et~al.(2002)Lodhi, Saunders, Shawe-Taylor, Cristianini, and
  Watkins}]{LodhiSSCW02}
Huma Lodhi, Craig Saunders, John Shawe-Taylor, Nello Cristianini, and
  Christopher J. C.~H. Watkins. 2002.
\newblock Text classification using string kernels.
\newblock \emph{Journal of Machine Learning Research}, 2:419--444.

\bibitem[{Malmasi and Zampieri(2017)}]{malmasi-VarDial-2017}
Shervin Malmasi and Marcos Zampieri. 2017.
\newblock {Arabic Dialect Identification Using iVectors and ASR Transcripts}.
\newblock In \emph{Proceedings of the VarDial Workshop of EACL}, pages
  178--183.

\bibitem[{McClosky et~al.(2006)McClosky, Charniak, and
  Johnson}]{McClosky-NAACL-2006}
David McClosky, Eugene Charniak, and Mark Johnson. 2006.
\newblock {Effective Self-training for Parsing}.
\newblock In \emph{Proceedings of NAACL}, pages 152--159.

\bibitem[{Pan et~al.(2010)Pan, Ni, Sun, Yang, and Chen}]{Pan-WWW-2010}
Sinno~Jialin Pan, Xiaochuan Ni, Jian-Tao Sun, Qiang Yang, and Zheng Chen. 2010.
\newblock {Cross-domain Sentiment Classification via Spectral Feature
  Alignment}.
\newblock In \emph{Proceedings of WWW}, pages 751--760.

\bibitem[{Popescu and Grozea(2012)}]{PopescuG12}
Marius Popescu and Cristian Grozea. 2012.
\newblock Kernel methods and string kernels for authorship analysis.
\newblock In \emph{Proceedings of CLEF (Online Working Notes/Labs/Workshop)}.

\bibitem[{Popescu et~al.(2017)Popescu, Grozea, and Ionescu}]{marius-KES-2017}
Marius Popescu, Cristian Grozea, and Radu~Tudor Ionescu. 2017.
\newblock {HASKER: An efficient algorithm for string kernels. Application to
  polarity classification in various languages}.
\newblock In \emph{Proceedings of KES}, pages 1755--1763.

\bibitem[{Popescu and Ionescu(2013)}]{popescu-ionescu:2013:BEA8}
Marius Popescu and Radu~Tudor Ionescu. 2013.
\newblock {The Story of the Characters, the DNA and the Native Language}.
\newblock In \emph{Proceedings of the Eighth Workshop on Innovative Use of NLP
  for Building Educational Applications}, pages 270--278.

\bibitem[{Shawe-Taylor and Cristianini(2004)}]{taylor-Cristianini-cup-2004}
John Shawe-Taylor and Nello Cristianini. 2004.
\newblock \emph{{Kernel Methods for Pattern Analysis}}.
\newblock Cambridge University Press.

\bibitem[{Sun et~al.(2016)Sun, Feng, and Saenko}]{Sun-AAAI-2016}
Baochen Sun, Jiashi Feng, and Kate Saenko. 2016.
\newblock {Return of Frustratingly Easy Domain Adaptation}.
\newblock In \emph{Proceedings of AAAI}, pages 2058--2065.

\bibitem[{Zampieri et~al.(2017)Zampieri, Malmasi, Ljube\v{s}i\'{c}, Nakov, Ali,
  Tiedemann, Scherrer, and Aepli}]{dsl2017}
Marcos Zampieri, Shervin Malmasi, Nikola Ljube\v{s}i\'{c}, Preslav Nakov, Ahmed
  Ali, J\"{o}rg Tiedemann, Yves Scherrer, and No\"{e}mi Aepli. 2017.
\newblock {Findings of the VarDial Evaluation Campaign 2017}.
\newblock In \emph{Proceedings of VarDial Workshop of EACL}, pages 1--15.

\end{thebibliography}
\bibliographystyle{acl_natbib}

\end{document}